\documentclass[runningheads]{llncs}

\usepackage[T1]{fontenc}
\usepackage[utf8]{inputenc}
\usepackage{graphicx,verbatim}
\usepackage{booktabs}
\usepackage{multirow}
\usepackage{amsmath, amssymb, amsfonts}
\usepackage{mathtools}
\usepackage{lmodern}
\usepackage{microtype}
\usepackage{xcolor}
\usepackage{caption}
\usepackage{subcaption}
\usepackage{float}
\usepackage[breakable, skins]{tcolorbox}
\usepackage{tikz}
\usetikzlibrary{arrows.meta, positioning, calc, fit, shapes.geometric,
                decorations.pathreplacing, patterns, shadows.blur, backgrounds}

\usepackage[hidelinks]{hyperref}

\definecolor{coolA}{HTML}{1F3D4F}
\definecolor{coolAm}{HTML}{2F6F84}
\definecolor{coolAl}{HTML}{8DB7C2}
\definecolor{coolAll}{HTML}{D6E5EA}

\definecolor{warmB}{HTML}{C24B2A}
\definecolor{warmBm}{HTML}{E07B5A}
\definecolor{warmBl}{HTML}{F1B49F}
\definecolor{warmBll}{HTML}{FAE3D8}

\definecolor{gold}{HTML}{C19A2A}
\definecolor{goldl}{HTML}{F1D77C}

\definecolor{sage}{HTML}{627A6B}
\definecolor{sagel}{HTML}{B0C2B5}

\definecolor{inkdark}{HTML}{1B1F2A}
\definecolor{inkmed}{HTML}{4C5263}
\definecolor{inklight}{HTML}{A8AEBC}
\definecolor{bgwarm}{HTML}{FBF7EE}

\tikzset{>=Triangle}
\tikzset{font=\sffamily}
\tikzset{every node/.style={inner sep=0pt, outer sep=0pt}}

\tikzset{arrowMain/.style={->, line width=0.85pt, color=inkmed}}
\tikzset{arrowGhost/.style={->, densely dotted, line width=0.55pt, color=inkmed!70}}
\tikzset{arrowFlow/.style={->, line width=1.6pt, color=coolA!85, double, double distance=1pt}}

\tikzset{imgframe/.style={draw=inkmed!55, line width=0.5pt, inner sep=1pt}}
\tikzset{scattercanvas/.style={draw=inkmed!35, line width=0.6pt, fill=bgwarm!50, rounded corners=14pt, inner sep=6pt}}

\tikzset{chipcool/.style={circle, draw=coolA!70, fill=coolA!22, minimum size=12pt, inner sep=0, font=\scriptsize\sffamily\bfseries\color{coolA!30!black}}}
\tikzset{chipwarm/.style={circle, draw=warmB!70, fill=warmB!22, minimum size=12pt, inner sep=0, font=\scriptsize\sffamily\bfseries\color{warmB!30!black}}}

\tikzset{encoderbox/.style={shape=regular polygon, regular polygon sides=6, draw=coolA!80, line width=0.7pt, fill=coolA!10, inner sep=2pt, align=center, font=\small\bfseries\sffamily\color{coolA!30!black}, minimum size=2.8cm}}

\tikzset{bankblob/.style={draw=inkmed!30, line width=0.55pt, dashed, fill=inkmed!4, rounded corners=18pt}}

\newcommand{\scatterdot}[2]{\node[circle, draw=#2!90, fill=#2!55, minimum size=4.6pt, inner sep=0, line width=0.4pt] at #1 {};}

\newcommand{\layerstack}[3]{%
  \foreach \i in {0,...,\numexpr#1-1\relax} {
    \pgfmathsetmacro{\yy}{\i*0.30}
    \pgfmathtruncatemacro{\op}{25+\i*7}
    \fill[#2!\op, rounded corners=2pt] (0, \yy) rectangle (#3, \yy+0.24);
    \draw[#2!70, very thin, rounded corners=2pt] (0, \yy) rectangle (#3, \yy+0.24);
  }
}

\newcommand\samethanks[1][\value{footnote}]{\footnotemark[#1]}

\begin{document}

\title{Transition-Aware best-of-$N$ sampling for Longitudinal
       Chest X-ray Reports}


\author{H.~Ibrahim Gulluk\inst{1}\fnmsep\thanks{Equal contribution.} \and
        Max Van Puyvelde\inst{2,3}\samethanks \and
        Wim Van Criekinge\inst{3} \and
        Olivier Gevaert\inst{2}}
\authorrunning{H.~I. Gulluk et al.}
\institute{Department of Electrical Engineering, Stanford University,
        Stanford, CA, USA\\
        \email{gulluk@stanford.edu} \and
        Department of Biomedical Data Science,
        Stanford University School of Medicine, Stanford, CA, USA\\
        \email{\{maxvpuyv,ogevaert\}@stanford.edu} \and
        Department of Mathematical Modelling, Statistics \& Bioinformatics,
        Ghent University, Ghent, Belgium\\
        \email{wim.vancriekinge@ugent.be}}

\maketitle

\begin{abstract}
In longitudinal clinical practice, every chest X-ray is read in the
context of the patient's \emph{prior} exam, and much of what the
radiologist communicates is the change from one visit to the next. To
the best of our knowledge, we present the first \emph{training-free}
best-of-$N$ sampling scheme for pre-trained chest X-ray report
generators that is explicitly aware of this longitudinal
prior$\rightarrow$current transition. We call it
\textbf{transition-aware best-of-$N$ sampling}: each report is
split into sentences and embedded into an unordered set in
$\mathbb{R}^{d}$; each (prior, current) pair is reduced to a fixed-dim
\emph{directional} vector via a set-to-set distance designed to encode
the change between the two sets; and candidates are scored by cosine
distance from their candidate transition vector to a cached bank of
ground-truth training transition vectors, aggregated as \textsc{min} or
\textsc{kNN}. We instantiate the framework with four
directional set distances (mean-shift, novelty residual,
directed-Hausdorff anchor, and cost-weighted optimal transport) and
evaluate on a multi-visit AP/PA cohort, running inference under three
prompts on three vision--language generators. Transition-aware
best-of-$N$ outperforms random selection across the board, with the
largest relative gains on the Impression section.
\end{abstract}

\section{Introduction}
\label{sec:intro}

Chest radiography is the most frequently performed imaging modality in
medicine, and automating chest X-ray report generation can substantially
ease radiologist workload while improving reporting consistency. Modern
vision--language models (VLMs) can already draft credible reports
\cite{liu2019clinically,li2023dynamic,endo2021retrieval,bannur2406maira,chen2024chexagent},
and a simple training-free way to extract further quality from such
generators is \emph{best-of-$N$ sampling}: drawing $N$ candidate reports
per image and selecting the one that is the best in 
terms if the preferences or the performance.

The vast majority of recent chest X-ray report generators -- and the
corresponding best-of-$N$ pipelines built on top of them -- treat each
test image in isolation, scoring candidates against a distribution of
single-image reference reports
\cite{gulluk2026sdr,bannur2406maira,chen2024chexagent}. In real clinical
practice, however, chest X-rays are rarely read in isolation: most exams
are acquired as follow-ups, and the radiologist's report is largely
organised around how the patient has changed since the prior study
(``stable interval'', ``new opacity'', ``improved aeration'').
Modelling this longitudinal context has been shown to improve generation
quality when fed to the decoder
\cite{bannur2023learning,nicolson2024longitudinal}, but to our knowledge
the same idea has not yet been exploited at the selection stage of a
best-of-$N$ pipeline. We argue that for a longitudinal image the
appropriate notion of best response should be selected based on the 
transitions from the prior visit to the current visit. As some transitions 
might not be clinically valid although those can be generated by a language model.

We turn this intuition into a concrete pipeline. Each report is split
into sentences and embedded into an unordered set in $\mathbb{R}^{d}$ by
a frozen sentence transformer
(Sec.~\ref{sec:report-rep}, Fig.~\ref{fig:report-encoding}). A
\emph{directional} set-to-set distance then reduces each (prior, current)
pair to a single fixed-dim vector
(Sec.~\ref{sec:transition-rep}, Fig.~\ref{fig:method-overview}~(a)). A
bank of such ground-truth training transition vectors is cached once. At
test time, each candidate report becomes its own transition vector and
is cosine-scored against the bank
(Fig.~\ref{fig:method-overview}~(b)). Crucially, an individually
plausible-looking candidate report can nonetheless describe a
clinically implausible change from the patient's prior study --
e.g.\ silently dropping a chronic finding or hallucinating an acute
worsening that the image does not support. Scoring in transition space
penalises exactly these candidates, because their displacement from the
prior departs from the displacements observed in training; report-space
scoring, by contrast, only checks that the candidate looks like
some real report. The transition bank therefore provides a prior
on \emph{how reports change between visits}, not just on what individual
reports look like. We instantiate this idea with a small family of
directional set distances, and benchmark it on a multi-visit cohort
drawn from a publicly available chest X-ray dataset across several
vision--language generators and prompting regimes; transition-aware
best-of-$N$ outperforms random selection on a wide range of text-overlap
and clinical-content metrics.

\section{Related work}
\label{sec:related}

Early systems combined a CNN encoder with an LSTM/Transformer decoder
\cite{liu2019clinically,li2023dynamic} or retrieved canned sentences
from a memory of training reports
\cite{endo2021retrieval}. More recent work pairs domain-specific visual
encoders with general-purpose LLMs: R2GenGPT
\cite{wang2023r2gengpt} aligns a Swin-Transformer visual feature stream
to a frozen LLaMA-2 via a learned projection; CheXagent
\cite{chen2024chexagent} instruction-tunes a clinical LLM jointly with
a CXR vision encoder on a curated mixture of 28 CXR datasets;
MAIRA-2 \cite{bannur2406maira} couples the RAD-DINO image
encoder with Vicuna-7B and a per-finding grounding head to produce
spatially grounded reports.

Several recent works exploit the patient's prior study at
generation time. BioViL-T \cite{bannur2023learning}
self-supervises a CNN+ViT encoder that explicitly fuses current and
prior CXR pairs and is used for temporal classification, phrase
grounding and longitudinal report generation. The authors
\cite{nicolson2024longitudinal} condition the report decoder on the
prior study (handling its absence with a learned placeholder) and add a
CXR-BERT semantic-similarity reward during fine-tuning. Both target the
decoder; in contrast we operate downstream at the best-of-$N$ selection
stage and require no changes to the generator.

Representing a report as an unordered set of sentence embeddings is a
natural fit for chest X-ray reading: standard set-to-set distances --
Chamfer, Hausdorff, optimal transport \cite{villani2009optimal},
Hungarian matching \cite{kuhn1955hungarian} -- yield continuous,
permutation-invariant scores between two reports. The closest existing
recipe to ours is SDR \cite{gulluk2026sdr}, which uses such
sentence-set distances both as a GRPO reward and as the scorer for
single-image best-of-$N$ selection; it remains image-independent and
does not encode change between visits.

Best-of-$N$ sampling with a learned verifier is a standard test-time
scaling tool in the LLM literature
\cite{cobbe2021training,lightman2024let}: $N$ candidates are drawn from
a stochastic generator and the highest-scoring one is kept. Our
pipeline has the same shape but the scorer operates in
transition space rather than at the candidate level.

\section{Method}
\label{sec:method}

\subsection{Sentence-set report representation}
\label{sec:report-rep}

%
\begin{figure}[t]
    \centering
    \resizebox{\linewidth}{!}{%
    \begin{tikzpicture}[
        sentbox/.style={draw=warmBm, rounded corners=2pt, line width=0.6pt,
                        fill=warmBl!55, inner sep=3pt, text width=5.8cm,
                        align=left, font=\footnotesize\color{inkdark}},
        sectbar/.style={line width=2.2pt, line cap=round},
        sectlbl/.style={font=\footnotesize\bfseries\sffamily,
                        inner sep=0pt, anchor=base west},
        stbox/.style={draw=gold!75, rounded corners=5pt, line width=0.8pt,
                      fill=gold!10, align=center, minimum width=2.7cm,
                      inner xsep=6pt, inner ysep=8pt},
        sttitle/.style={font=\normalsize\bfseries\sffamily\color{gold!25!black},
                        align=center, inner sep=0pt},
        stsub/.style={font=\scriptsize\ttfamily\color{inkmed}, inner sep=0pt},
        vecbox/.style={draw=coolAm, rounded corners=1.5pt, line width=0.6pt,
                       fill=coolAl!55, minimum width=0.42cm, minimum height=1.40cm,
                       inner sep=0},
        veclbl/.style={font=\footnotesize\sffamily\color{inkdark}, inner sep=1pt},
        xraycap/.style={font=\footnotesize\sffamily\color{inkmed}, inner sep=1pt},
        frozenpill/.style={draw=gold!70, rounded corners=2pt, fill=gold!18,
                           font=\tiny\bfseries\sffamily\color{gold!30!black},
                           inner xsep=3pt, inner ysep=1.5pt},
        arw/.style={->, >=Triangle, line width=0.9pt, color=inkmed!75},
    ]

    \coordinate (fOrigin) at (0, 0);
    \node[sectlbl, color=coolA!35!black, anchor=base west]
         at (fOrigin) (fhdr) {Findings};

    \node[sentbox, anchor=north west] at ([yshift=-5pt] fhdr.south west) (f1)
        {Stable position of right IJ tunneled hemodialysis catheter.};
    \node[sentbox, anchor=north west] at ([yshift=-3pt] f1.south west) (f2)
        {Catheter tip projects over the distal SVC.};
    \node[sentbox, anchor=north west] at ([yshift=-3pt] f2.south west) (f3)
        {Cardiac and mediastinal contours are within normal limits.};
    \node[sentbox, anchor=north west] at ([yshift=-3pt] f3.south west) (f4)
        {The lungs are clear.};
    \node[sentbox, anchor=north west] at ([yshift=-3pt] f4.south west) (f5)
        {No acute osseous abnormality.};

    \draw[sectbar, color=coolA!60]
        ([xshift=-6pt, yshift=4pt] fhdr.base west) --
        ([xshift=-6pt] f5.south west);

    \node[sectlbl, color=warmB!40!black, anchor=base west]
         at ([yshift=-14pt] f5.south west) (ihdr) {Impression};

    \node[sentbox, anchor=north west] at ([yshift=-5pt] ihdr.south west) (i1)
        {Right IJ tunneled hemodialysis catheter with the tip overlying the distal SVC.};
    \node[sentbox, anchor=north west] at ([yshift=-3pt] i1.south west) (i2)
        {No evidence of acute cardiopulmonary process.};

    \draw[sectbar, color=warmB!65]
        ([xshift=-6pt, yshift=4pt] ihdr.base west) --
        ([xshift=-6pt] i2.south west);

    \node[imgframe, anchor=east] (xray)
        at ($(f1.north west)!0.5!(i2.south west) + (-0.80cm,0)$)
        {\includegraphics[width=3.8cm]{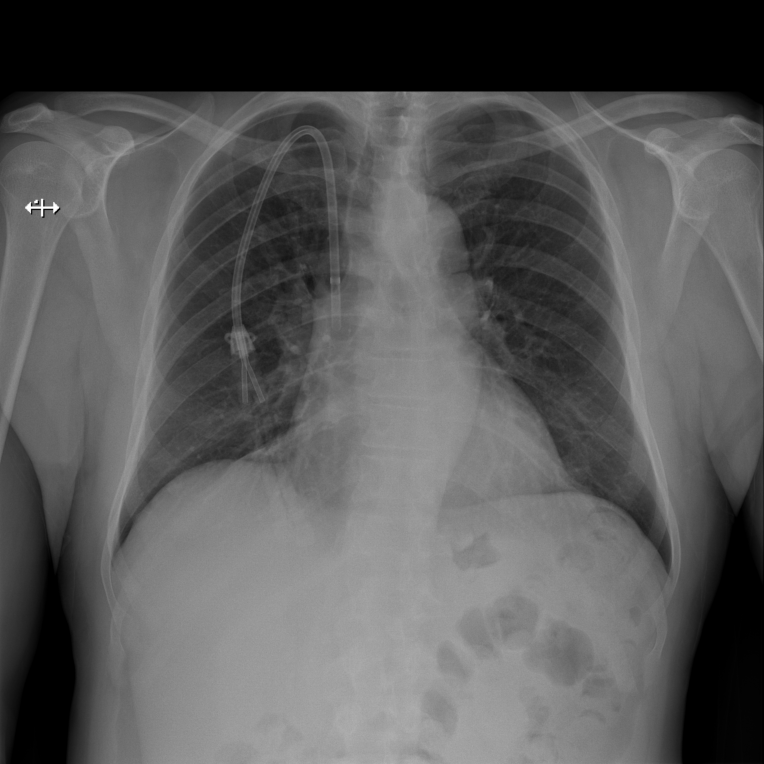}};
    \node[xraycap, anchor=north] at ([yshift=-5pt] xray.south) {Chest X-ray};

    \node[stbox, minimum height=8.9cm] (st)
        at ($(f1.east)!0.5!(i2.east) + (3.1cm,0)$) {};

    \node[sttitle, anchor=north] at ([yshift=-0.35cm] st.north)
        {Sentence\\[-1pt]Transformer};

    \node[frozenpill, anchor=north]
          at ([yshift=-1.15cm] st.north) {FROZEN};

    \begin{scope}[shift={($(st.center) + (-0.65cm,-0.30cm)$)}]
        \layerstack{5}{gold}{1.30}
    \end{scope}

    \node[stsub, anchor=south] at ([yshift=0.30cm] st.south)
        {all-mpnet-base-v2};

    \foreach \s in {f1,f2,f3,f4,f5,i1,i2}{
        \draw[arw] (\s.east) -- (\s.east -| st.west);
    }

    \foreach \s/\idx/\sec in {%
        f1/1/F, f2/2/F, f3/3/F, f4/4/F, f5/5/F,
        i1/1/I, i2/2/I%
    }{
        \node[vecbox] (v\s) at ([xshift=2.0cm] st.east |- \s) {};
        \foreach \k in {1,2,3,4,5,6}{
            \draw[coolA!55, very thin]
                ([yshift=-0.2cm*\k] v\s.north west) --
                ([yshift=-0.2cm*\k] v\s.north east);
        }
        \draw[arw] (st.east |- \s) -- (v\s.west);
        \node[veclbl, right=3pt of v\s] {$\mathbf{e}^{\sec}_{\idx}$};
    }

    \coordinate (setBraceTop) at ($(vf1.north east) + (1.15cm,0)$);
    \coordinate (setBraceBot) at (setBraceTop |- vi2.south east);

    \draw[decorate, decoration={brace, amplitude=5pt}, line width=0.7pt,
          color=inkmed!85]
         (setBraceTop) -- (setBraceBot)
         node[midway, right=9pt, align=left,
              font=\footnotesize\sffamily\color{inkdark}]
         {$\mathcal{E}(r)$\\[3pt]
          $\subset \mathbb{R}^{d}$};

    \end{tikzpicture}%
    }
    \caption{\textbf{Sentence-level encoding of a chest X-ray report.}
    Each visit pairs a radiograph with a free-text report composed of a
    Findings and an Impression section. We split both sections into
    individual sentences and embed each sentence independently with the
    frozen pre-trained \texttt{all-mpnet-base-v2} sentence transformer,
    producing one $d$-dimensional vector per sentence. The resulting
    unordered collection of sentence embeddings
    $\mathcal{E}(r)=\{\mathbf{e}^{F}_{1},\dots,\mathbf{e}^{F}_{5},
    \mathbf{e}^{I}_{1},\mathbf{e}^{I}_{2}\}\subset\mathbb{R}^{d}$ serves
    as the report representation throughout Sec.~\ref{sec:method}.}
    \label{fig:report-encoding}
\end{figure}
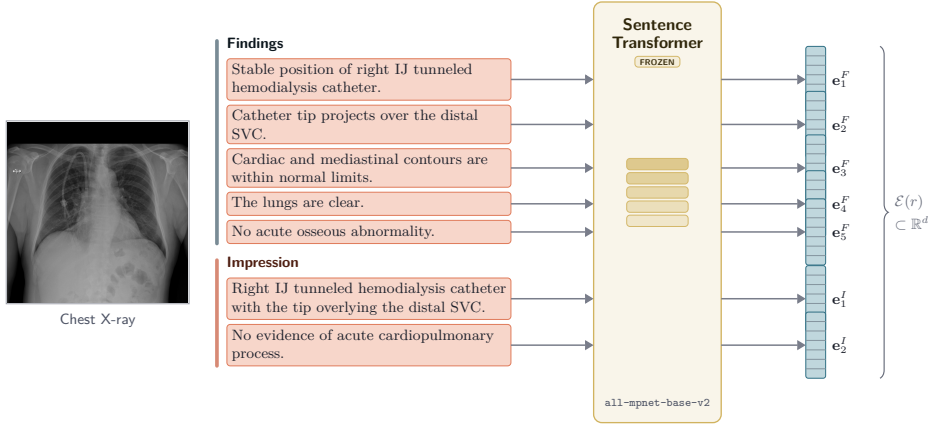

We adopt the notation of the SDR formulation~\cite{gulluk2026sdr}: a
chest X-ray report $r$ consists of a Findings section $r^{F}$ and an
Impression section $r^{I}$. We split each section into individual
sentences using a standard sentence segmenter, yielding
$r^{F}=(s^{F}_{1},\dots,s^{F}_{n_F})$ and
$r^{I}=(s^{I}_{1},\dots,s^{I}_{n_I})$ with sentence counts
$n_F, n_I\in\mathbb{N}$ that vary across studies. Each sentence $s$ is
mapped to a fixed-dimensional embedding
$\mathbf{e}=E_{\phi}(s)\in\mathbb{R}^{d}$ by a frozen pre-trained
sentence transformer $E_{\phi}$ (specifically
\texttt{all-mpnet-base-v2} \cite{reimers2019sentence}, $d=768$).
The report is represented by the two unordered embedding sets
\begin{equation}
    \mathcal{E}^{F}(r) \;=\; \bigl\{\,E_{\phi}(s^{F}_{i})\,:\,1\le i\le n_F\,\bigr\},
    \qquad
    \mathcal{E}^{I}(r) \;=\; \bigl\{\,E_{\phi}(s^{I}_{j})\,:\,1\le j\le n_I\,\bigr\},
    \label{eq:report-sets}
\end{equation}
both subsets of $\mathbb{R}^{d}$ (Fig.~\ref{fig:report-encoding}). Being
sets, they are invariant to the order of the underlying sentences, which
matches the observation that the listing order of individual findings
carries no diagnostic meaning. Throughout the rest of this section the
unhatted $r$ denotes a ground-truth report and $\hat y$ denotes one
candidate report produced by a generator $\pi$.

\subsection{Transition representation}
\label{sec:transition-rep}

%
\begin{figure}[t]
    \centering
    \begin{subfigure}[b]{0.46\linewidth}
        \centering
        \resizebox{\linewidth}{!}{%
        \begin{tikzpicture}[
            visitlbl/.style={font=\footnotesize\itshape\rmfamily\color{inkmed},
                             inner sep=2pt},
            xraycap/.style={font=\scriptsize\itshape\color{inkmed},
                             inner sep=1pt},
            setlbl/.style={font=\small\rmfamily\color{inkdark},
                            inner sep=2pt},
            tvecarrow/.style={->, >=Triangle,
                                line width=2pt, color=gold!85!black},
            tveclbl/.style={font=\footnotesize\rmfamily\color{gold!30!black},
                             inner sep=4pt, fill=bgwarm, rounded corners=2pt,
                             draw=gold!70, line width=0.5pt},
        ]
            \node[imgframe, anchor=south]
                 (pxray) at (-2.9, 1.95)
                 {\includegraphics[height=2.4cm]{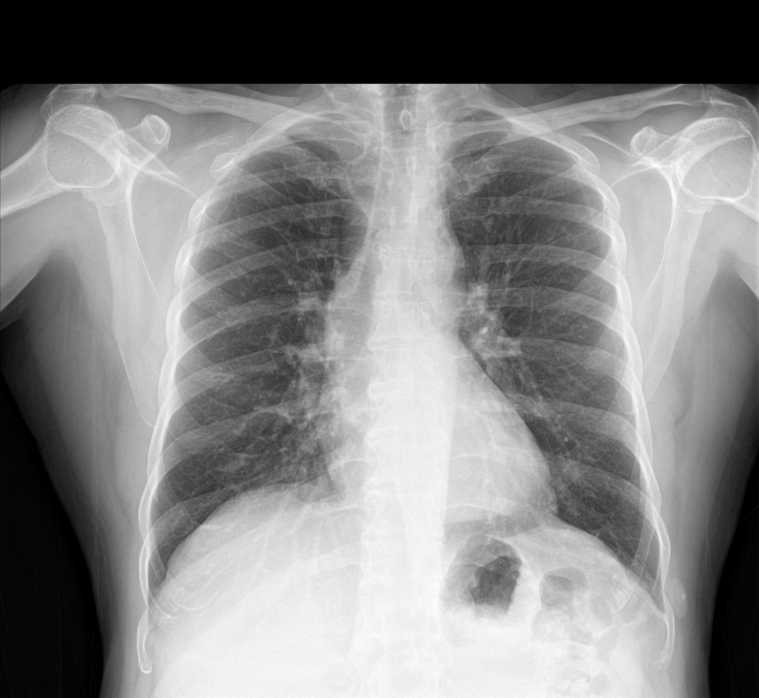}};

            \node[bankblob, minimum width=2.6cm, minimum height=2.0cm,
                   draw=coolA!50, fill=coolA!8] (pblob)
                  at (-2.9, 0) {};
            \node[setlbl, color=coolA!30!black, anchor=south]
                 at ([yshift=2pt] pblob.north)
                 {$\mathcal{E}^{S}\bigl(r^{(k-1)}_{p}\bigr)$};
            \node[visitlbl, anchor=north] at ([yshift=-3pt] pblob.south)
                 {prior visit};
            \foreach \x/\y in {-3.5/0.4, -3.1/-0.3, -2.6/0.5, -2.3/-0.2,
                                -2.8/0.0, -3.3/-0.1}{
                \scatterdot{(\x, \y)}{coolA}
            }

            \node[imgframe, anchor=south]
                 (cxray) at (2.9, 1.95)
                 {\includegraphics[height=2.4cm]{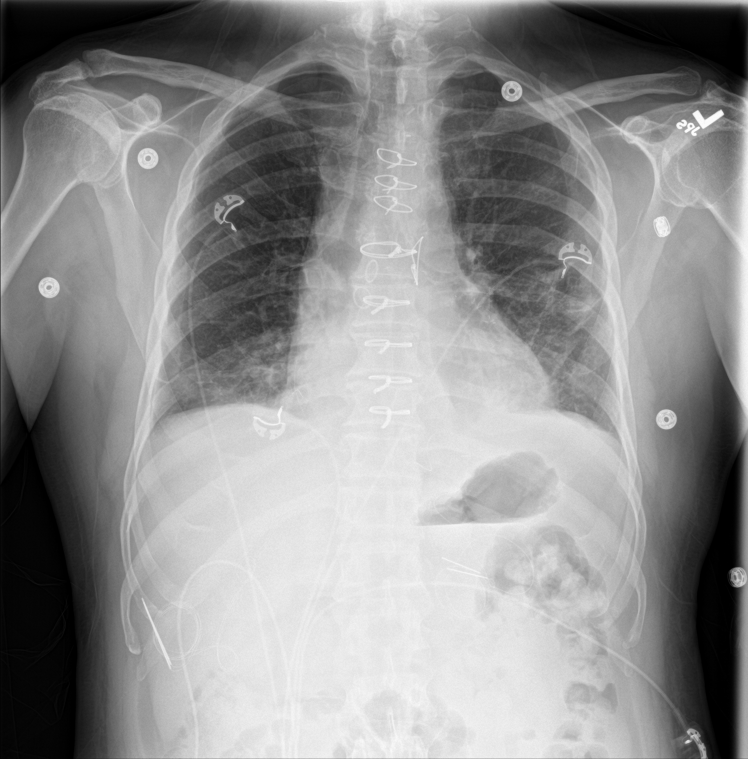}};

            \node[bankblob, minimum width=2.6cm, minimum height=2.0cm,
                   draw=warmB!55, fill=warmB!8] (cblob)
                  at (2.9, 0) {};
            \node[setlbl, color=warmB!30!black, anchor=south]
                 at ([yshift=2pt] cblob.north)
                 {$\mathcal{E}^{S}\bigl(r^{(k)}_{p}\bigr)$};
            \node[visitlbl, anchor=north] at ([yshift=-3pt] cblob.south)
                 {current visit};
            \foreach \x/\y in {2.4/0.3, 3.0/0.5, 3.5/-0.1, 2.7/-0.4,
                                3.3/0.1}{
                \scatterdot{(\x, \y)}{warmB}
            }

            \draw[tvecarrow] ([xshift=2pt] pblob.east) --
                              ([xshift=-2pt] cblob.west);

            \node[tveclbl, anchor=south]
                 at ($(pblob.east)!0.5!(cblob.west) + (0, 0.32)$)
                 {$g^{(k)} = \mathcal{T}(\mathcal{A},\mathcal{B}) \in \mathbb{R}^{d'}$};

        \end{tikzpicture}
        }
        \caption{Transition computation. A directional set distance
        $\mathcal{T}$ reduces the (prior, current) sentence-embedding
        sets to a single vector $g^{(k)}\in\mathbb{R}^{d'}$.}
        \label{fig:transition-compute}
    \end{subfigure}\hfill
    \begin{subfigure}[b]{0.52\linewidth}
        \centering
        \resizebox{\linewidth}{!}{%
        \begin{tikzpicture}[
            bank/.style={shape=regular polygon, regular polygon sides=4,
                          draw=sage!75, fill=sage!40,
                          minimum size=3pt, inner sep=0, line width=0.3pt},
            bankNN/.style={shape=regular polygon, regular polygon sides=4,
                            draw=sage!90, fill=sage!70,
                            minimum size=4.6pt, inner sep=0, line width=0.4pt},
            cand/.style={shape=regular polygon, regular polygon sides=4,
                          draw=warmB!85, fill=warmB!55,
                          minimum size=7.5pt, inner sep=0, line width=0.7pt,
                          rotate=45},
            sel/.style={shape=regular polygon, regular polygon sides=6,
                         draw=gold!80!black, fill=gold!75,
                         minimum size=12pt, inner sep=0, line width=0.9pt},
            distline/.style={densely dotted, color=inkmed!70,
                              line width=0.55pt,
                              shorten <=2pt, shorten >=2pt},
            selline/.style={color=gold!75!black, line width=1.5pt,
                             shorten <=2pt, shorten >=2pt},
            glabel/.style={font=\scriptsize\rmfamily\color{warmB!30!black},
                            inner sep=1pt},
            slabel/.style={font=\footnotesize\bfseries\rmfamily\color{gold!25!black},
                            inner sep=1.5pt},
            panel/.style={draw=inkmed!25, fill=bgwarm!50,
                           rounded corners=4pt, line width=0.5pt},
            spacelbl/.style={font=\scriptsize\itshape\sffamily\color{inkmed},
                              inner sep=3pt},
            ringA/.style={draw=sage!55, fill=none, line width=0.6pt, dashed},
            ringB/.style={draw=sage!35, fill=none, line width=0.5pt, densely dotted},
            legend/.style={font=\scriptsize\rmfamily\color{inkdark},
                            inner sep=2pt},
        ]

            \node[panel, minimum width=9.6cm, minimum height=5.2cm] (P)
                at (-1.6, 0) {};
            \node[spacelbl, anchor=north west]
                 at ([xshift=6pt, yshift=-6pt] P.north west)
                 {Transition space $\mathbb{R}^{d'}$};

            \draw[ringB] (-0.4, 0.0) ellipse (2.8 and 2.0);
            \draw[ringA] (-0.4, 0.0) ellipse (2.0 and 1.45);

            \foreach \x/\y in {
                -0.7/-0.2,  -0.2/0.5,  -1.2/0.4,  0.0/0.9,  -0.4/0.0,
                -1.6/0.7,   0.3/0.6,  -0.8/1.1,  -0.3/0.2, -1.1/0.1,
                -0.9/0.8,   0.1/0.2,  -1.4/0.0, -0.6/0.6, -0.1/1.2,
                -1.0/0.3,  -0.4/0.8, -1.3/1.0, -0.2/-0.2,-1.7/0.3,
                -0.7/0.4,   0.2/0.7,  -0.9/-0.1,-0.5/1.0, -1.5/0.8,
                 0.0/0.3,  -1.1/0.6, -0.3/1.1, -0.8/0.2,  0.3/1.0,
                -1.2/-0.2, -0.1/0.6, -0.6/0.0, -1.6/0.5, -0.2/0.4,
                -1.0/1.2,   0.1/0.8, -1.4/0.2, -0.5/0.1, -0.7/0.9,
                -1.8/0.6,   0.4/0.4, -0.3/-0.4,-1.3/-0.4,
                 0.4/-0.3, -1.5/-0.2, 0.2/-0.4
            }{
                \node[bank] at (\x,\y) {};
            }

            \coordinate (nn1) at (-1.80,  0.60);
            \coordinate (nn2) at (-1.70,  0.30);
            \coordinate (nn3) at (-1.30, -0.40);
            \coordinate (nn4) at (-1.60,  0.70);
            \coordinate (nns) at (-1.40,  0.20);

            \node[cand] (g1) at (-4.9,  1.5) {};
            \node[glabel, above=1pt of g1] {$\hat g^{(1)}$};

            \node[cand] (g2) at (-5.1, -0.4) {};
            \node[glabel, left=1pt of g2] {$\hat g^{(2)}$};

            \node[cand] (g3) at (-4.4, -1.7) {};
            \node[glabel, below=1pt of g3] {$\hat g^{(3)}$};

            \node[cand] (g4) at (-3.7,  1.1) {};
            \node[glabel, above=1pt of g4] {$\hat g^{(4)}$};

            \node[sel] (gs) at (-2.95, 0.15) {};
            \node[slabel, below=4pt of gs] {$\hat g^{\star}$};

            \node[bankNN] at (nn1) {};
            \node[bankNN] at (nn2) {};
            \node[bankNN] at (nn3) {};
            \node[bankNN] at (nn4) {};
            \node[bankNN] at (nns) {};

            \draw[distline] (g1) -- (nn1);
            \draw[distline] (g2) -- (nn2);
            \draw[distline] (g3) -- (nn3);
            \draw[distline] (g4) -- (nn4);
            \draw[selline]  (gs) -- (nns);

            \node[bank, anchor=west] (lt0)
                  at ([xshift=8pt, yshift=-0.55cm] P.south west) {};
            \node[legend, right=5pt of lt0] {training transition $g_{t}$};

            \node[cand, anchor=west] (lc0)
                  at ([xshift=3.6cm] lt0.west) {};
            \node[legend, right=5pt of lc0] {candidate $\hat g^{(k)}$};

            \node[sel, anchor=west] (ls0)
                  at ([xshift=6.5cm] lt0.west) {};
            \node[legend, right=5pt of ls0] {selected $\hat g^{\star}$};

        \end{tikzpicture}%
        }
        \caption{Selection in transition space. Each gray square is a
        training transition; orange diamonds are candidate transitions
        $\hat g^{(k)}$; the gold hexagon is the selected response.}
        \label{fig:transition-select}
    \end{subfigure}
    \caption{\textbf{Transition representation and transition-aware
    best-of-$N$ sampling.} \textbf{(a)} A directional set distance
    $\mathcal{T}$ reduces the (prior, current) sentence-embedding sets to a
    single vector $g^{(k)}\in\mathbb{R}^{d'}$. \textbf{(b)} At test time,
    each of the $N$ candidate reports becomes its own transition vector
    $\hat g^{(k)}$ and is scored by cosine distance to a cached bank of
    $T$ training transition vectors (Eq.~\eqref{eq:per-bank-dist}); the
    chosen response $\hat g^{\star}$ is the one closest to the bank under
    the chosen aggregation
    (Eqs.~\eqref{eq:agg}--\eqref{eq:selection-rule}).}
    \label{fig:method-overview}
\end{figure}

The patient-level cohort consists of multi-visit studies. For a patient
$p$ with visits $r^{(1)}_{p}, r^{(2)}_{p}, \dots$ ordered by study
date, a \textbf{transition} is the ordered pair
$\bigl(r^{(k-1)}_{p},\, r^{(k)}_{p}\bigr)$ with $k\ge 2$. Each transition
is represented in the same sentence-set space as in Eq.~\eqref{eq:report-sets},
once per section $S\in\{F,I\}$, by the two embedding sets
$\mathcal{E}^{S}\bigl(r^{(k-1)}_{p}\bigr)$ and $\mathcal{E}^{S}\bigl(r^{(k)}_{p}\bigr)$.

\paragraph{Directional set distance.}
A \emph{directional set distance} is any map
\[
    \mathcal{T} \,:\;
    2^{\mathbb{R}^d} \times 2^{\mathbb{R}^d}
    \;\longrightarrow\;
    \mathbb{R}^{d'}
\]
that takes a pair of finite, non-empty embedding sets
$\mathcal{A}=\{\mathbf{a}_i\}_{i=1}^{n}$, $\mathcal{B}=\{\mathbf{b}_j\}_{j=1}^{m}$
and returns a vector of fixed dimension $d'$, where $d'$ may depend on
the choice of $\mathcal{T}$ but not on $n,m$. Unlike the symmetric scalar
set-to-set distances commonly used for best-of-$N$ scoring, $\mathcal{T}$
is asymmetric ($\mathcal{T}(\mathcal{A},\mathcal{B})$ encodes a change
from $\mathcal{A}$ to $\mathcal{B}$), and its vector-valued
output enables cosine-style comparisons across distinct
$(\mathcal{A},\mathcal{B})$ pairs.

We instantiate four concrete designs, summarised in
Table~\ref{tab:vectors} (Appendix~\ref{sec:vectors-appendix}). Let
$d(\mathbf{u},\mathbf{v})=\tfrac{1}{2}\bigl(1-\mathbf{u}^{\top}\mathbf{v}\bigr)$
be the cosine distance between two unit-norm embeddings, and write
$\mu(\mathcal{X})=\frac{1}{|\mathcal{X}|}\sum_{\mathbf{x}\in\mathcal{X}}\mathbf{x}$,
$\mathrm{NN}_{\mathcal{A}}(\mathbf{b})=\arg\min_{\mathbf{a}\in\mathcal{A}}d(\mathbf{b},\mathbf{a})$.

\begin{description}
    \item[\textsc{Mean-Shift}:] The centroid shift of $\mathcal{B}$ relative
        to $\mathcal{A}$:
        \begin{equation}
            \mathcal{T}_{\textsc{ms}}(\mathcal{A},\mathcal{B})
            \;=\;
            \mu(\mathcal{B}) - \mu(\mathcal{A}) \;\in\; \mathbb{R}^{d}.
            \label{eq:ms}
        \end{equation}
    \item[\textsc{Novelty}:] The mean residual of each current-side sentence
        relative to its nearest neighbour in the prior set:
        \begin{equation}
            \mathcal{T}_{\textsc{nov}}(\mathcal{A},\mathcal{B})
            \;=\;
            \frac{1}{|\mathcal{B}|}\sum_{\mathbf{b}\in\mathcal{B}}
            \Bigl(\mathbf{b} - \mathrm{NN}_{\mathcal{A}}(\mathbf{b})\Bigr)
            \;\in\; \mathbb{R}^{d}.
            \label{eq:nov}
        \end{equation}
        This is the forward-only Chamfer direction, which captures what
        is new in $\mathcal{B}$ relative to $\mathcal{A}$.
    \item[\textsc{Dir-Hausdorff}:] The displacement at the directed-Hausdorff
        anchor -- the single worst-covered point in $\mathcal{B}$
        relative to $\mathcal{A}$, paired with its nearest neighbour:
        \begin{equation}
            \mathbf{b}^{\star}
            \;=\;
            \operatorname*{arg\,max}_{\mathbf{b}\in\mathcal{B}}\,
            \min_{\mathbf{a}\in\mathcal{A}} d(\mathbf{b},\mathbf{a}),
            \quad
            \mathcal{T}_{\textsc{dh}}(\mathcal{A},\mathcal{B})
            \;=\;
            \mathbf{b}^{\star}-\mathrm{NN}_{\mathcal{A}}(\mathbf{b}^{\star}).
            \label{eq:dh}
        \end{equation}
    \item[\textsc{Cost-OT}:] A cost-weighted optimal-transport
        displacement. Let $M_{ij}=d(\mathbf{a}_i,\mathbf{b}_j)$ be the
        cosine cost matrix and let $\gamma^{*}\in\mathbb{R}^{n\times m}_{\ge 0}$
        be the optimal transport plan between the uniform marginals
        $\tfrac{1}{n}\mathbf{1}_n$ and $\tfrac{1}{m}\mathbf{1}_m$
        \cite{villani2009optimal}. We define
        \begin{equation}
            \mathcal{T}_{\textsc{cot}}(\mathcal{A},\mathcal{B})
            \;=\;
            \sum_{i=1}^{n}\sum_{j=1}^{m}
            \gamma^{*}_{ij}\,M_{ij}\,(\mathbf{b}_j-\mathbf{a}_i)
            \;\in\; \mathbb{R}^{d}.
            \label{eq:cot}
        \end{equation}
        Because every per-pair displacement is re-weighted by both transport
        mass and cost, the trivial collapse to $\mu(\mathcal{B})-\mu(\mathcal{A})$
        that mass-only weighting would produce is avoided.
\end{description}

For each section $S\in\{F,I\}$ the transition is then represented by the
section-level vector
\begin{equation}
    g^{S}\!\bigl(r^{(k-1)}_{p},\,r^{(k)}_{p}\bigr)
    \;=\;
    \mathcal{T}\!\Bigl(
        \mathcal{E}^{S}\bigl(r^{(k-1)}_{p}\bigr),\;
        \mathcal{E}^{S}\bigl(r^{(k)}_{p}\bigr)
    \Bigr)
    \;\in\;\mathbb{R}^{d'}.
    \label{eq:transition-vec}
\end{equation}

\subsection{Transition-aware best-of-$N$ sampling}
\label{sec:bestofN}

Let $\mathcal{T}^{\,\mathrm{train}} = \{(r^{(t-1)},\, r^{(t)})\}_{t=1}^{N}$
be the corpus of training transitions. For each $t$ and each section
$S\in\{F,I\}$ we pre-compute the ground-truth training transition vector
\begin{equation}
    g_{t}^{S}
    \;=\;
    \mathcal{T}\!\bigl(\mathcal{E}^{S}(r^{(t-1)}),\,\mathcal{E}^{S}(r^{(t)})\bigr)
    \;\in\;\mathbb{R}^{d'},
    \label{eq:bank-vec}
\end{equation}
yielding two banks
$\mathcal{B}^{S}_{\mathcal{T}}=\{g_{t}^{S}\}_{t=1}^{N}$ that depend
only on the frozen encoder $E_{\phi}$ and the directional design
$\mathcal{T}$. The banks are cached once per training corpus.

\paragraph{Candidate scoring.}
For a multi-visit test patient $p$, at visit $k\ge 2$, the generator
$\pi$ produces $K$ candidate reports
$\hat y^{(k,1)}, \dots, \hat y^{(k,K)} \stackrel{\text{i.i.d.}}{\sim}\pi(\cdot\mid x^{(k)})$.
For each candidate we extract the Findings and Impression sections from
$\hat y^{(k,j)}$, embed them into sets, and compute the section-level
\emph{candidate transition vector}
\begin{equation}
    \hat g^{S,j}
    \;=\;
    \mathcal{T}\!\Bigl(
        \mathcal{E}^{S}\bigl(r^{(k-1)}_{p}\bigr),\;
        \mathcal{E}^{S}\bigl(\hat y^{(k,j)}\bigr)
    \Bigr)
    \;\in\;\mathbb{R}^{d'}.
    \label{eq:cand-vec}
\end{equation}
Each candidate is then scored by its cosine distance to every entry of the
section bank,
\begin{equation}
    \mathfrak{D}_{S,\,t}\bigl(\hat y^{(k,j)}\bigr)
    \;=\;
    \tfrac{1}{2}\!\left(
        1 - \frac{\langle \hat g^{S,j},\, g_{t}^{S}\rangle}
                  {\Vert \hat g^{S,j}\Vert_{2}\,\Vert g_{t}^{S}\Vert_{2}}
    \right),
    \qquad t=1,\dots,N,
    \label{eq:per-bank-dist}
\end{equation}
and the per-section bank distance is taken as one of two aggregations:
\begin{equation}
    \mathfrak{D}^{\min}_{S}(\hat y)
    \;=\;
    \min_{1\le t\le N} \mathfrak{D}_{S,\,t}(\hat y),
    \qquad
    \mathfrak{D}^{k\mathrm{NN}}_{S}(\hat y)
    \;=\;
    \frac{1}{k}\!\sum_{t\,\in\,\mathcal{N}_{k}(\hat y)}\!\mathfrak{D}_{S,\,t}(\hat y),
    \label{eq:agg}
\end{equation}
where $\mathcal{N}_{k}(\hat y)\subseteq\{1,\dots,N\}$ indexes the $k$
training transitions with the smallest $\mathfrak{D}_{S,\,t}(\hat y)$.
$\mathfrak{D}^{\min}$ asks whether the candidate's transition resembles
any single training transition, while $\mathfrak{D}^{k\mathrm{NN}}$ is a
noise-robust soft alternative.

\paragraph{Selection rule.}
The total bank distance for a candidate is the sum across the two
sections,
\begin{equation}
    \mathfrak{D}(\hat y)
    \;=\;
    \mathfrak{D}_{F}(\hat y) \;+\; \mathfrak{D}_{I}(\hat y),
    \label{eq:total-dist}
\end{equation}
and the selected response is the candidate with the smallest total
distance,
\begin{equation}
    \hat y^{\star}
    \;=\;
    \operatorname*{arg\,min}_{j\in\{1,\dots,K\}}\, \mathfrak{D}\!\bigl(\hat y^{(k,j)}\bigr).
    \label{eq:selection-rule}
\end{equation}
Figure~\ref{fig:method-overview} illustrates the full pipeline: the
candidate's transition vector $\hat g^{(k)}$ is constructed from the
patient's prior and the generated current, then scored against the
cached training bank of transition vectors.

\section{Experimental setup}
\label{sec:setup}

\subsection{Multi-visit cohort}
\label{sec:cohort}

We derive a multi-visit AP/PA cohort from ReXGradient-160K
\cite{zhang2025rexgradient}, keeping patients with at least two qualifying exams,
sampling a single view (AP or PA) per patient, and forming one
transition per consecutive visit pair. The 22{,}745 training transitions
populate the bank $\mathcal{B}^{S}_{\mathcal{T}}$; the 1{,}727 test
transitions are the evaluation set. Per-split counts are in
Appendix~\ref{sec:cohort-appendix}.

\subsection{Candidate generation}
\label{sec:generation}

We evaluate three vision--language models -- Gemini-2.5-Flash-Lite,
Gemini-3.1-Flash-Lite-preview, and Mistral-Small-2603 -- under three
prompts: \textbf{P1} (zero-shot, image-only), \textbf{P2} (few-shot with
five random training reports as Findings/Impression examples), and
\textbf{P3} (zero-shot conditioned on the patient's prior ground-truth
report; the only prompt that places longitudinal context in the
generator's input). Full prompt texts are in
Appendix~\ref{sec:prompts-appendix}. For each (model, prompt)
configuration and each test transition we sample $K=5$ candidate reports
with distinct seeds.

\subsection{Evaluation metrics}
\label{sec:metrics}

Selected responses are scored against the current-visit ground-truth
report with the standard combination of NLP overlap metrics and
clinical-content metrics, separately for Findings and Impressions; the
full list of metrics and references is given in
Appendix~\ref{sec:metrics-appendix}. The random baseline is a
uniform pick over the five candidates, averaged over five seeds. 

\section{Results}
\label{sec:results}

\sloppy

\subsection{Headline metrics averaged across runs}
\label{sec:results-headline}

Tables~\ref{tab:avg-impressions} and~\ref{tab:avg-findings} report the
seven headline metrics on Impressions and Findings, averaged across all
9 (model, prompt) configurations. The best transition row per column is
in bold; subscripts give the relative change versus random.

\begin{table}[t]
  \caption{\textbf{Impressions} -- headline metrics averaged across the runs in which every listed method has data. Subscripts give the relative change vs.\ the \emph{random baseline} row at the bottom (\textit{italicised}). The best transition row per column is in \textbf{bold}. The very last row reports the per-metric mean across all transition methods.}
  \label{tab:avg-impressions}
  \centering
  \setlength{\tabcolsep}{3pt}
  \resizebox{\textwidth}{!}{
  \begin{tabular}{lccccccc}
    \toprule
    Method & BLEU-1 & ROUGE-1 & ROUGE-L & METEOR & BERTSc & RadGraph & CheX-14 \\
    \midrule
    \textsc{Mean-Shift} / \textsc{min} & 0.163\textsubscript{(+3.0\%)} & 0.153\textsubscript{(+6.4\%)} & 0.137\textsubscript{(+8.2\%)} & 0.159\textsubscript{(+2.8\%)} & \textbf{0.174\textsubscript{(+3.3\%)}} & \textbf{0.064\textsubscript{(+13.2\%)}} & \textbf{0.440\textsubscript{(+3.4\%)}} \\
    \textsc{Mean-Shift} / \textsc{knn} & 0.163\textsubscript{(+2.8\%)} & 0.152\textsubscript{(+6.1\%)} & 0.136\textsubscript{(+7.9\%)} & 0.159\textsubscript{(+2.7\%)} & 0.172\textsubscript{(+2.5\%)} & 0.064\textsubscript{(+12.2\%)} & 0.438\textsubscript{(+3.0\%)} \\
    \textsc{Novelty} / \textsc{min} & 0.165\textsubscript{(+3.8\%)} & 0.153\textsubscript{(+6.6\%)} & 0.137\textsubscript{(+8.4\%)} & 0.161\textsubscript{(+3.8\%)} & 0.173\textsubscript{(+2.9\%)} & 0.062\textsubscript{(+10.1\%)} & 0.435\textsubscript{(+2.2\%)} \\
    \textsc{Novelty} / \textsc{knn} & 0.165\textsubscript{(+4.2\%)} & \textbf{0.153\textsubscript{(+6.9\%)}} & \textbf{0.137\textsubscript{(+8.6\%)}} & 0.161\textsubscript{(+4.1\%)} & 0.173\textsubscript{(+2.7\%)} & 0.063\textsubscript{(+11.0\%)} & 0.433\textsubscript{(+1.6\%)} \\
    \textsc{Dir-Hausdorff} / \textsc{min} & 0.166\textsubscript{(+4.8\%)} & 0.153\textsubscript{(+6.7\%)} & 0.136\textsubscript{(+7.8\%)} & 0.162\textsubscript{(+4.3\%)} & 0.174\textsubscript{(+3.3\%)} & 0.062\textsubscript{(+9.7\%)} & 0.430\textsubscript{(+1.0\%)} \\
    \textsc{Dir-Hausdorff} / \textsc{knn} & \textbf{0.167\textsubscript{(+5.0\%)}} & 0.153\textsubscript{(+6.7\%)} & 0.136\textsubscript{(+7.9\%)} & \textbf{0.162\textsubscript{(+4.4\%)}} & 0.173\textsubscript{(+3.2\%)} & 0.062\textsubscript{(+10.0\%)} & 0.430\textsubscript{(+1.0\%)} \\
    \textsc{Cost-OT} / \textsc{min} & 0.163\textsubscript{(+3.0\%)} & 0.152\textsubscript{(+6.3\%)} & 0.137\textsubscript{(+8.0\%)} & 0.160\textsubscript{(+3.2\%)} & 0.173\textsubscript{(+3.1\%)} & 0.064\textsubscript{(+12.2\%)} & 0.438\textsubscript{(+2.9\%)} \\
    \textsc{Cost-OT} / \textsc{knn} & 0.164\textsubscript{(+3.1\%)} & 0.153\textsubscript{(+6.4\%)} & 0.136\textsubscript{(+7.9\%)} & 0.160\textsubscript{(+3.2\%)} & 0.173\textsubscript{(+2.9\%)} & 0.064\textsubscript{(+12.0\%)} & 0.437\textsubscript{(+2.7\%)} \\
    \midrule
    \textit{Random baseline (reference)} & \textit{0.159} & \textit{0.143} & \textit{0.126} & \textit{0.155} & \textit{0.168} & \textit{0.057} & \textit{0.426} \\
    \midrule
    \textbf{Mean of transition methods} & \textbf{0.164}\textsubscript{(+3.7\%)} & \textbf{0.153}\textsubscript{(+6.5\%)} & \textbf{0.137}\textsubscript{(+8.1\%)} & \textbf{0.160}\textsubscript{(+3.6\%)} & \textbf{0.173}\textsubscript{(+3.0\%)} & \textbf{0.063}\textsubscript{(+11.3\%)} & \textbf{0.435}\textsubscript{(+2.2\%)} \\
    \bottomrule
  \end{tabular}
  }
\end{table}
\begin{table}[t]
  \caption{\textbf{Findings} -- headline metrics averaged across the runs in which every listed method has data. Subscripts give the relative change vs.\ the \emph{random baseline} row at the bottom (\textit{italicised}). The best transition row per column is in \textbf{bold}. The very last row reports the per-metric mean across all transition methods.}
  \label{tab:avg-findings}
  \centering
  \setlength{\tabcolsep}{3pt}
  \resizebox{\textwidth}{!}{
  \begin{tabular}{lccccccc}
    \toprule
    Method & BLEU-1 & ROUGE-1 & ROUGE-L & METEOR & BERTSc & RadGraph & CheX-14 \\
    \midrule
    \textsc{Mean-Shift} / \textsc{min} & 0.269\textsubscript{(-0.0\%)} & 0.288\textsubscript{(+0.7\%)} & 0.194\textsubscript{(+1.4\%)} & 0.248\textsubscript{(+0.2\%)} & 0.201\textsubscript{(+2.5\%)} & 0.105\textsubscript{(+4.5\%)} & 0.376\textsubscript{(+2.1\%)} \\
    \textsc{Mean-Shift} / \textsc{knn} & 0.269\textsubscript{(-0.0\%)} & 0.288\textsubscript{(+0.7\%)} & 0.194\textsubscript{(+1.4\%)} & 0.248\textsubscript{(+0.3\%)} & 0.201\textsubscript{(+2.4\%)} & 0.105\textsubscript{(+4.6\%)} & \textbf{0.377\textsubscript{(+2.4\%)}} \\
    \textsc{Novelty} / \textsc{min} & 0.270\textsubscript{(+0.2\%)} & 0.288\textsubscript{(+0.7\%)} & 0.194\textsubscript{(+1.3\%)} & 0.249\textsubscript{(+0.6\%)} & 0.200\textsubscript{(+1.9\%)} & 0.104\textsubscript{(+3.8\%)} & 0.375\textsubscript{(+1.7\%)} \\
    \textsc{Novelty} / \textsc{knn} & 0.269\textsubscript{(+0.0\%)} & 0.288\textsubscript{(+0.4\%)} & 0.193\textsubscript{(+1.0\%)} & 0.248\textsubscript{(+0.3\%)} & 0.199\textsubscript{(+1.6\%)} & 0.104\textsubscript{(+3.0\%)} & 0.374\textsubscript{(+1.5\%)} \\
    \textsc{Dir-Hausdorff} / \textsc{min} & 0.270\textsubscript{(+0.2\%)} & 0.287\textsubscript{(+0.3\%)} & 0.193\textsubscript{(+1.1\%)} & 0.248\textsubscript{(+0.4\%)} & 0.200\textsubscript{(+2.0\%)} & 0.104\textsubscript{(+3.0\%)} & 0.374\textsubscript{(+1.5\%)} \\
    \textsc{Dir-Hausdorff} / \textsc{knn} & \textbf{0.270\textsubscript{(+0.4\%)}} & 0.288\textsubscript{(+0.5\%)} & 0.194\textsubscript{(+1.2\%)} & \textbf{0.249\textsubscript{(+0.6\%)}} & 0.200\textsubscript{(+2.1\%)} & 0.104\textsubscript{(+3.0\%)} & 0.374\textsubscript{(+1.6\%)} \\
    \textsc{Cost-OT} / \textsc{min} & 0.269\textsubscript{(+0.1\%)} & \textbf{0.288\textsubscript{(+0.7\%)}} & \textbf{0.194\textsubscript{(+1.5\%)}} & 0.248\textsubscript{(+0.4\%)} & \textbf{0.201\textsubscript{(+2.5\%)}} & \textbf{0.105\textsubscript{(+4.7\%)}} & 0.375\textsubscript{(+1.9\%)} \\
    \textsc{Cost-OT} / \textsc{knn} & 0.269\textsubscript{(+0.0\%)} & 0.288\textsubscript{(+0.6\%)} & 0.194\textsubscript{(+1.4\%)} & 0.248\textsubscript{(+0.3\%)} & 0.201\textsubscript{(+2.4\%)} & 0.105\textsubscript{(+4.5\%)} & 0.375\textsubscript{(+1.8\%)} \\
    \midrule
    \textit{Random baseline (reference)} & \textit{0.269} & \textit{0.286} & \textit{0.191} & \textit{0.247} & \textit{0.196} & \textit{0.101} & \textit{0.368} \\
    \midrule
    \textbf{Mean of transition methods} & \textbf{0.269}\textsubscript{(+0.1\%)} & \textbf{0.288}\textsubscript{(+0.6\%)} & \textbf{0.194}\textsubscript{(+1.3\%)} & \textbf{0.248}\textsubscript{(+0.4\%)} & \textbf{0.201}\textsubscript{(+2.2\%)} & \textbf{0.105}\textsubscript{(+3.9\%)} & \textbf{0.375}\textsubscript{(+1.8\%)} \\
    \bottomrule
  \end{tabular}
  }
\end{table}

\paragraph{Impressions.}
Transition-aware best-of-$N$ delivers substantial relative gains on
Impressions: the strongest configurations give
$+8.5\%$ ROUGE-L (\textsc{Novelty}/\textsc{kNN}),
$+13.6\%$ RadGraph averaged-F1 (\textsc{Mean-Shift}/\textsc{min}), and
$+4.4\%$ METEOR (\textsc{Dir-Hausdorff}/\textsc{kNN}). Impressions are
short, change-oriented summary statements -- e.g.\ ``stable since
prior'', ``new opacity in the right lower lobe'' -- and match the regime
in which a transition signal should carry the most ranking information.

\paragraph{Findings.}
Every \textsc{min} and \textsc{kNN} aggregation of every transition
vector beats random on every headline metric. The strongest result on
BERTScore-F1 ($+3.1\%$) and RadGraph averaged-F1 ($+5.1\%$) is from
\textsc{Cost-OT}/\textsc{min}; the best CheXbert F-14 is from
\textsc{Mean-Shift}/\textsc{kNN} ($+2.2\%$).

\subsection{Per-model and per-prompt breakdowns}
\label{sec:results-per-run-pointer}

Appendices~\ref{sec:per-run-appendix} and~\ref{sec:delta-appendix}
break the averaged numbers down by (model, prompt). Transition-aware
best-of-$N$ beats random on nearly every Impressions configuration and
on the majority of Findings ones, with the largest gains on the weaker
prompts and a consistent trend across both Gemini and Mistral.

\section{Conclusion}
\label{sec:conclusion}

We introduced \emph{transition-aware best-of-$N$ sampling} for chest
X-ray report generation. Each (prior, current) report pair is reduced
to a single fixed-dim vector by a directional set-to-set distance, and
candidates are scored at test time against a cached bank of
ground-truth training transition vectors. To our knowledge this is the
first training-free best-of-$N$ scheme for pre-trained chest X-ray
report generators that explicitly conditions the scorer on the
patient's prior exam. Across nine (model, prompt) configurations on a
multi-visit AP/PA cohort the pipeline consistently beats random
selection on both report sections, with the largest relative gains on
the change-oriented Impressions and a stable trend across both Gemini
and Mistral generators.

Several extensions are worth pursuing: learnt directional encoders
trained to maximise alignment with a target clinical metric, larger
candidate pools $N$, longer histories beyond a single prior visit, and
combining \emph{prior-conditioned generation} with
\emph{prior-conditioned selection} as two complementary ways of
injecting longitudinal context.

\clearpage
\bibliographystyle{splncs04}
\bibliography{refs}

\clearpage
\appendix
\section{Prompts}
\label{sec:prompts-appendix}

We reproduce verbatim the three prompts used in the experiments. The
few-shot examples in P2 are sampled at random from the training
corpus for every test query; the placeholders \{prior\_findings\} and
\{prior\_impression\} in P3 are filled at query time with the patient's
prior ground-truth Findings and Impression respectively.

\subsection*{P1 -- Zero-shot, image-only}

\begin{tcolorbox}[breakable, enhanced, colback=bgwarm!50,
                   colframe=inkmed!40, sharp corners, boxrule=0.4pt,
                   fonttitle=\bfseries, title=\texttt{prompt1.txt}]
\ttfamily\small
You are a radiology report generation model specialized in chest X-rays.\\[2pt]
Generate a concise clinical report for the given image.\\[2pt]
Strict requirements:\\
- Output ONLY in this exact format:\\
\phantom{xxx}Findings: <text>\\
\phantom{xxx}Impression: <text>\\
- Do NOT include explanations, disclaimers, or any extra text.\\
- Do NOT include phrases such as ``consult a doctor'' or ``this is not medical advice''.\\
- Use professional radiology language.\\
- Keep it concise and structured.\\[2pt]
Output:
\end{tcolorbox}

\subsection*{P2 -- Few-shot with random training examples}

\begin{tcolorbox}[breakable, enhanced, colback=bgwarm!50,
                   colframe=inkmed!40, sharp corners, boxrule=0.4pt,
                   fonttitle=\bfseries, title=\texttt{prompt2.txt}]
\ttfamily\small
You are a radiology report generation model. Given a chest X-ray image, generate a concise radiology report.\\[2pt]
Follow these strict rules:\\
- Output ONLY in this format:\\
\phantom{xxx}Findings: <text>\\
\phantom{xxx}Impression: <text>\\
- Do NOT include any explanations, disclaimers, or additional commentary.\\
- Do NOT say things like ``consult a doctor'' or ``this is not medical advice''.\\
- Match the writing style, tone, and structure of the examples below.\\
- Be concise and clinically accurate.\\[2pt]
Here are example reports:\\[2pt]
Example 0:\\
Findings: Mild cardiomegaly. No edema. No consolidation or effusion. No pneumothorax.\\
Impression: Mild cardiomegaly\\[2pt]
Example 1:\\
Findings: No pneumonia is seen. Minimal peribronchial thickening is noted. The heart is within normal limits in size. No bony abnormality is seen.\\
Impression: No pneumonia. Mild peribronchial thickening.\\[2pt]
Example 2:\\
Findings: The heart size and mediastinal contours are within normal limits. Both lungs are clear. The visualized skeletal structures are unremarkable.\\
Impression: No active cardiopulmonary disease.\\[2pt]
Example 3:\\
Findings: The lungs are well-expanded. The interstitial markings are increased bilaterally. Patchy areas of confluence are noted in the mid to lower left lung and at the right lung base. The heart and pulmonary vascularity are normal. The mediastinum is normal in width. There is multilevel degenerative disc disease of the thoracic spine.\\
Impression: Bilateral interstitial pneumonia with patchy areas of alveolar infiltrate. No pulmonary edema. No pleural effusion. Followup PA and lateral chest X-ray is recommended in 3-4 weeks following trial of antibiotic therapy to ensure resolution and exclude underlying malignancy.\\[2pt]
Example 4:\\
Findings: The heart size and mediastinal contours are within normal limits. Both lungs are clear. The visualized skeletal structures are unremarkable.\\
Impression: No active disease.\\[2pt]
Now generate the report for the given chest X-ray image.\\[2pt]
Output:
\end{tcolorbox}

\subsection*{P3 -- Zero-shot with prior report as context}

\begin{tcolorbox}[breakable, enhanced, colback=bgwarm!50,
                   colframe=inkmed!40, sharp corners, boxrule=0.4pt,
                   fonttitle=\bfseries, title=\texttt{prompt3.txt}]
\ttfamily\small
You are a radiology report generation model specialized in chest X-rays.\\[2pt]
You are given a chest X-ray image of a patient who has at least one prior chest X-ray exam. Use the patient's prior report below as clinical context, and then generate the report for the new image -- describing what is present now, including any changes (new findings, worsening, improvement, resolution) relative to the prior.\\[2pt]
Patient's prior chest X-ray report:\\
Findings: \{prior\_findings\}\\
Impression: \{prior\_impression\}\\[2pt]
Strict requirements:\\
- Output ONLY in this exact format:\\
\phantom{xxx}Findings: <text>\\
\phantom{xxx}Impression: <text>\\
- Do NOT include explanations, disclaimers, or any extra text.\\
- Do NOT include phrases such as ``consult a doctor'' or ``this is not medical advice''.\\
- Do NOT copy the prior report verbatim -- describe the new image, noting changes.\\
- Use professional radiology language.\\
- Keep it concise and structured.\\[2pt]
Output:
\end{tcolorbox}
\clearpage
\section{Directional set distances}
\label{sec:vectors-appendix}

Table~\ref{tab:vectors} compactly summarises the four directional
set-to-set distances $\mathcal{T}$ defined in
Sec.~\ref{sec:transition-rep} (\textsc{Mean-Shift}, \textsc{Novelty},
\textsc{Dir-Hausdorff}, \textsc{Cost-OT}). Each maps a pair of
sentence-embedding sets to a single fixed-dim vector that can be
cosine-compared across transitions.

\begin{table}[t]
  \captionsetup{font=small, labelfont={bf,small}}
  \caption{Directional set distances used in this work.
  Each maps two sentence-embedding sets $\mathcal{A}, \mathcal{B}\subset\mathbb{R}^d$
  to a single fixed-dim vector that can be cosine-compared across transitions.
  $\mu(\cdot)$ denotes the mean of the set;
  $\mathrm{NN}_{\mathcal{A}}(\mathbf{b})$ the cosine-nearest neighbour of
  $\mathbf{b}$ in $\mathcal{A}$; and $\gamma^{*}$ the optimal-transport plan
  between the uniform measures on $\mathcal{A}$ and $\mathcal{B}$ with cost
  matrix $M_{ij}=d(\mathbf{a}_i, \mathbf{b}_j)$. Output dim is $d$ for all four.}
  \label{tab:vectors}
  \centering
  \footnotesize
  \setlength{\tabcolsep}{4pt}
  \begin{tabular}{ll}
    \toprule
    Name & Vector \\
    \midrule
    \textsc{Mean-Shift}     & $\mathcal{T}_{\textsc{ms}}(\mathcal{A},\mathcal{B}) = \mu(\mathcal{B})-\mu(\mathcal{A})$ \\
    \textsc{Novelty}        & $\mathcal{T}_{\textsc{nov}}(\mathcal{A},\mathcal{B}) = \frac{1}{m}\sum_{\mathbf{b}\in\mathcal{B}}\bigl(\mathbf{b}-\mathrm{NN}_{\mathcal{A}}(\mathbf{b})\bigr)$ \\
    \textsc{Dir-Hausdorff}  & $\mathcal{T}_{\textsc{dh}}(\mathcal{A},\mathcal{B}) = \mathbf{b}^{\star}-\mathrm{NN}_{\mathcal{A}}(\mathbf{b}^{\star}),\;\; \mathbf{b}^{\star}{=}\arg\max_{\mathbf{b}\in\mathcal{B}}\min_{\mathbf{a}\in\mathcal{A}} d(\mathbf{b},\mathbf{a})$ \\
    \textsc{Cost-OT}        & $\mathcal{T}_{\textsc{cot}}(\mathcal{A},\mathcal{B}) = \sum_{i,j}\gamma^{*}_{ij}\,M_{ij}\,(\mathbf{b}_j-\mathbf{a}_i)$ \\
    \bottomrule
  \end{tabular}
\end{table}
\clearpage
\section{Multi-visit cohort statistics}
\label{sec:cohort-appendix}

Table~\ref{tab:cohort} reports per-split patient, visit and transition
counts for the multi-visit AP/PA cohort used throughout
Sec.~\ref{sec:cohort}.

\begin{table}[t]
  \captionsetup{font=small, labelfont={bf,small}}
  \caption{Multi-visit AP/PA chest X-ray cohort. Only patients with $\ge 2$
  qualifying exams are kept; one view (AP or PA) is chosen per patient.
  Transitions are the $(k{-}1, k)$ visit pairs that drive the
  transition bank and the test queries.}
  \label{tab:cohort}
  \centering
  \begin{tabular}{lrrr}
    \toprule
    Split & Patients & Visits & Transitions \\
    \midrule
    train & 19{,}492 & 42{,}237 & 22{,}745 \\
    valid &  1{,}411 &  3{,}090 &  1{,}679 \\
    test  &  1{,}459 &  3{,}186 &  1{,}727 \\
    \bottomrule
  \end{tabular}
\end{table}
\clearpage
\section{Evaluation metrics in detail}
\label{sec:metrics-appendix}

Selected responses are scored against the current-visit ground-truth
report with the following suite, separately for Findings and Impressions:

\begin{itemize}
    \item Sentence- and corpus-BLEU at $n=1{,}\dots{,}4$ \cite{papineni2002bleu}.
    \item ROUGE-1, ROUGE-2 and ROUGE-L, each as precision, recall and
        F-measure \cite{lin2004rouge}.
    \item METEOR \cite{banerjee2005meteor}.
    \item BERTScore-F1 with \texttt{roberta-large}, rescaled with
        baseline \cite{zhang2019bertscore}.
    \item COMET (\texttt{wmt22-comet-da}) \cite{rei2022comet}.
    \item chrF++ \cite{popovic2017chrf}.
    \item RadGraph entity, entity-relation and averaged F1
        \cite{delbrouck2022improving}.
    \item CheXbert F1 at 14 classes and 5 classes
        \cite{smit2020combining}.
\end{itemize}

The seven scores used in the headline tables (BLEU-1, ROUGE-1 F-measure,
ROUGE-L F-measure, METEOR, BERTScore-F1, RadGraph averaged-F1, CheXbert
F-14) are a representative subset spanning shallow text-overlap
(BLEU-1, ROUGE), soft lexical/semantic similarity
(METEOR, BERTScore-F1), and clinical-content correctness
(RadGraph, CheXbert).
\clearpage
\section{Per-(model, prompt) absolute values}
\label{sec:per-run-appendix}

Tables~\ref{tab:per-run-impressions} and \ref{tab:per-run-findings}
report the absolute headline values of the strongest single transition
method (\textsc{Novelty}/\textsc{min}) against the random baseline, one
row per (model, prompt) configuration. Three patterns hold across the
nine runs:

\begin{itemize}
    \item Transition-aware best-of-$N$ beats the matched random
        baseline on \emph{every} (model, prompt) configuration for
        ROUGE-L and RadGraph averaged-F1 on Impressions, and on the
        majority of configurations on Findings.
    \item The largest absolute gains appear on the weaker prompts P1
        and P2; under P3 (prior already in the prompt) the random
        baseline starts from a substantially higher absolute value and
        the method gain compresses but does not invert.
    \item The gain pattern is consistent across model families: both
        Gemini and Mistral see positive deltas, indicating that the
        transition signal is not tied to a single generator's failure
        modes.
\end{itemize}

\begin{table}[t]
  \caption{\textbf{Per-(model, prompt) absolute values on Impressions.}}
  \label{tab:per-run-impressions}
  \centering
  \setlength{\tabcolsep}{3pt}
  \resizebox{\textwidth}{!}{
  \begin{tabular}{llccccccc}
    \toprule
    Model & Prompt & BLEU-1 & ROUGE-1 & ROUGE-L & METEOR & BERTSc & RadGraph & CheX-14 \\
    \midrule
    \multicolumn{9}{l}{\textit{Random baseline (absolute values)}} \\
    Gemini-2.5-FL & P1 & 0.127 & 0.097 & 0.088 & 0.112 & 0.133 & 0.039 & 0.352 \\
    Gemini-2.5-FL & P2 & 0.152 & 0.133 & 0.123 & 0.139 & 0.150 & 0.056 & 0.374 \\
    Gemini-2.5-FL & P3 & 0.189 & 0.185 & 0.159 & 0.191 & 0.195 & 0.077 & 0.423 \\
    Gemini-3.1-FL & P1 & 0.161 & 0.136 & 0.116 & 0.151 & 0.156 & 0.052 & 0.417 \\
    Gemini-3.1-FL & P2 & 0.193 & 0.184 & 0.164 & 0.184 & 0.183 & 0.072 & 0.417 \\
    Gemini-3.1-FL & P3 & 0.206 & 0.206 & 0.171 & 0.221 & 0.201 & 0.079 & 0.439 \\
    Mistral-S-2603 & P1 & 0.079 & 0.037 & 0.034 & 0.070 & 0.131 & 0.017 & 0.434 \\
    Mistral-S-2603 & P2 & 0.131 & 0.119 & 0.117 & 0.124 & 0.163 & 0.049 & 0.528 \\
    Mistral-S-2603 & P3 & 0.188 & 0.193 & 0.165 & 0.201 & 0.200 & 0.071 & 0.444 \\
    \cmidrule(lr){1-9}
    \multicolumn{2}{l}{\textit{mean}} & 0.159 & 0.143 & 0.126 & 0.155 & 0.168 & 0.057 & 0.426 \\
    \midrule
    \multicolumn{9}{l}{\textit{\textsc{Novelty} / \textsc{min} (absolute values)}} \\
    Gemini-2.5-FL & P1 & 0.135\textsubscript{(+6.2\%)} & 0.110\textsubscript{(+13.4\%)} & 0.101\textsubscript{(+15.8\%)} & 0.120\textsubscript{(+7.1\%)} & 0.145\textsubscript{(+8.6\%)} & 0.047\textsubscript{(+21.8\%)} & 0.363\textsubscript{(+3.1\%)} \\
    Gemini-2.5-FL & P2 & 0.160\textsubscript{(+5.4\%)} & 0.151\textsubscript{(+13.6\%)} & 0.143\textsubscript{(+16.4\%)} & 0.150\textsubscript{(+7.7\%)} & 0.160\textsubscript{(+6.5\%)} & 0.066\textsubscript{(+19.0\%)} & 0.405\textsubscript{(+8.1\%)} \\
    Gemini-2.5-FL & P3 & 0.194\textsubscript{(+2.4\%)} & 0.192\textsubscript{(+3.6\%)} & 0.168\textsubscript{(+5.2\%)} & 0.195\textsubscript{(+2.0\%)} & 0.198\textsubscript{(+1.4\%)} & 0.078\textsubscript{(+1.5\%)} & 0.418\textsubscript{(-1.2\%)} \\
    Gemini-3.1-FL & P1 & 0.165\textsubscript{(+2.4\%)} & 0.145\textsubscript{(+6.6\%)} & 0.127\textsubscript{(+9.1\%)} & 0.156\textsubscript{(+3.2\%)} & 0.161\textsubscript{(+3.3\%)} & 0.054\textsubscript{(+5.4\%)} & 0.425\textsubscript{(+1.9\%)} \\
    Gemini-3.1-FL & P2 & 0.198\textsubscript{(+2.6\%)} & 0.191\textsubscript{(+3.9\%)} & 0.173\textsubscript{(+5.8\%)} & 0.186\textsubscript{(+0.8\%)} & 0.189\textsubscript{(+3.1\%)} & 0.079\textsubscript{(+10.0\%)} & 0.430\textsubscript{(+3.1\%)} \\
    Gemini-3.1-FL & P3 & 0.210\textsubscript{(+2.4\%)} & 0.213\textsubscript{(+3.3\%)} & 0.178\textsubscript{(+3.9\%)} & 0.224\textsubscript{(+1.4\%)} & 0.204\textsubscript{(+1.8\%)} & 0.081\textsubscript{(+2.5\%)} & 0.449\textsubscript{(+2.1\%)} \\
    Mistral-S-2603 & P1 & 0.079\textsubscript{(-0.9\%)} & 0.034\textsubscript{(-8.3\%)} & 0.032\textsubscript{(-5.9\%)} & 0.069\textsubscript{(-1.5\%)} & 0.135\textsubscript{(+3.5\%)} & 0.018\textsubscript{(+6.6\%)} & 0.451\textsubscript{(+3.9\%)} \\
    Mistral-S-2603 & P2 & 0.144\textsubscript{(+9.6\%)} & 0.139\textsubscript{(+17.5\%)} & 0.139\textsubscript{(+18.1\%)} & 0.139\textsubscript{(+11.9\%)} & 0.158\textsubscript{(-2.8\%)} & 0.065\textsubscript{(+31.8\%)} & 0.532\textsubscript{(+0.7\%)} \\
    Mistral-S-2603 & P3 & 0.196\textsubscript{(+4.2\%)} & 0.200\textsubscript{(+3.6\%)} & 0.173\textsubscript{(+4.8\%)} & 0.208\textsubscript{(+3.4\%)} & 0.206\textsubscript{(+2.7\%)} & 0.074\textsubscript{(+3.7\%)} & 0.443\textsubscript{(-0.3\%)} \\
    \cmidrule(lr){1-9}
    \multicolumn{2}{l}{\textit{mean}} & 0.165 & 0.153 & 0.137 & 0.161 & 0.173 & 0.062 & 0.435 \\
    \bottomrule
  \end{tabular}
  }
\end{table}
\begin{table}[t]
  \caption{\textbf{Per-(model, prompt) absolute values on Findings.} Strongest transition method (\textsc{Novelty} / \textsc{min}) vs.\ the random baseline; subscript is $\%$ change vs.\ random.}
  \label{tab:per-run-findings}
  \centering
  \setlength{\tabcolsep}{3pt}
  \resizebox{\textwidth}{!}{
  \begin{tabular}{llccccccc}
    \toprule
    Model & Prompt & BLEU-1 & ROUGE-1 & ROUGE-L & METEOR & BERTSc & RadGraph & CheX-14 \\
    \midrule
    \multicolumn{9}{l}{\textit{Random baseline (absolute values)}} \\
    Gemini-2.5-FL & P1 & 0.226 & 0.245 & 0.162 & 0.195 & 0.147 & 0.070 & 0.352 \\
    Gemini-2.5-FL & P2 & 0.254 & 0.277 & 0.188 & 0.223 & 0.182 & 0.092 & 0.357 \\
    Gemini-2.5-FL & P3 & 0.300 & 0.316 & 0.212 & 0.282 & 0.223 & 0.128 & 0.379 \\
    Gemini-3.1-FL & P1 & 0.293 & 0.306 & 0.198 & 0.277 & 0.188 & 0.097 & 0.375 \\
    Gemini-3.1-FL & P2 & 0.306 & 0.318 & 0.211 & 0.280 & 0.212 & 0.117 & 0.373 \\
    Gemini-3.1-FL & P3 & 0.299 & 0.327 & 0.212 & 0.312 & 0.227 & 0.126 & 0.372 \\
    Mistral-S-2603 & P1 & 0.217 & 0.228 & 0.150 & 0.182 & 0.160 & 0.052 & 0.341 \\
    Mistral-S-2603 & P2 & 0.234 & 0.254 & 0.179 & 0.204 & 0.200 & 0.096 & 0.428 \\
    Mistral-S-2603 & P3 & 0.292 & 0.307 & 0.211 & 0.270 & 0.228 & 0.129 & 0.338 \\
    \cmidrule(lr){1-9}
    \multicolumn{2}{l}{\textit{mean}} & 0.269 & 0.286 & 0.191 & 0.247 & 0.196 & 0.101 & 0.368 \\
    \midrule
    \multicolumn{9}{l}{\textit{\textsc{Novelty} / \textsc{min} (absolute values)}} \\
    Gemini-2.5-FL & P1 & 0.228\textsubscript{(+1.0\%)} & 0.249\textsubscript{(+1.9\%)} & 0.166\textsubscript{(+2.5\%)} & 0.200\textsubscript{(+2.1\%)} & 0.153\textsubscript{(+3.6\%)} & 0.076\textsubscript{(+8.2\%)} & 0.349\textsubscript{(-0.8\%)} \\
    Gemini-2.5-FL & P2 & 0.252\textsubscript{(-0.5\%)} & 0.279\textsubscript{(+0.9\%)} & 0.192\textsubscript{(+2.6\%)} & 0.223\textsubscript{(+0.0\%)} & 0.189\textsubscript{(+3.4\%)} & 0.097\textsubscript{(+5.3\%)} & 0.378\textsubscript{(+6.0\%)} \\
    Gemini-2.5-FL & P3 & 0.300\textsubscript{(+0.2\%)} & 0.317\textsubscript{(+0.3\%)} & 0.212\textsubscript{(+0.1\%)} & 0.282\textsubscript{(+0.2\%)} & 0.224\textsubscript{(+0.8\%)} & 0.128\textsubscript{(+0.1\%)} & 0.387\textsubscript{(+2.2\%)} \\
    Gemini-3.1-FL & P1 & 0.292\textsubscript{(-0.3\%)} & 0.305\textsubscript{(-0.3\%)} & 0.198\textsubscript{(-0.3\%)} & 0.276\textsubscript{(-0.4\%)} & 0.187\textsubscript{(-0.1\%)} & 0.095\textsubscript{(-1.8\%)} & 0.386\textsubscript{(+3.0\%)} \\
    Gemini-3.1-FL & P2 & 0.309\textsubscript{(+1.0\%)} & 0.321\textsubscript{(+1.1\%)} & 0.215\textsubscript{(+1.9\%)} & 0.283\textsubscript{(+1.1\%)} & 0.218\textsubscript{(+2.5\%)} & 0.123\textsubscript{(+5.3\%)} & 0.376\textsubscript{(+0.9\%)} \\
    Gemini-3.1-FL & P3 & 0.299\textsubscript{(+0.0\%)} & 0.327\textsubscript{(+0.1\%)} & 0.212\textsubscript{(-0.2\%)} & 0.312\textsubscript{(-0.1\%)} & 0.227\textsubscript{(+0.2\%)} & 0.124\textsubscript{(-1.5\%)} & 0.379\textsubscript{(+1.7\%)} \\
    Mistral-S-2603 & P1 & 0.222\textsubscript{(+2.3\%)} & 0.233\textsubscript{(+2.0\%)} & 0.153\textsubscript{(+1.8\%)} & 0.186\textsubscript{(+2.0\%)} & 0.168\textsubscript{(+4.9\%)} & 0.056\textsubscript{(+7.6\%)} & 0.349\textsubscript{(+2.1\%)} \\
    Mistral-S-2603 & P2 & 0.229\textsubscript{(-2.1\%)} & 0.255\textsubscript{(+0.3\%)} & 0.186\textsubscript{(+4.1\%)} & 0.206\textsubscript{(+1.1\%)} & 0.207\textsubscript{(+3.6\%)} & 0.112\textsubscript{(+16.2\%)} & 0.430\textsubscript{(+0.6\%)} \\
    Mistral-S-2603 & P3 & 0.294\textsubscript{(+0.6\%)} & 0.308\textsubscript{(+0.4\%)} & 0.211\textsubscript{(+0.1\%)} & 0.270\textsubscript{(+0.1\%)} & 0.229\textsubscript{(+0.2\%)} & 0.130\textsubscript{(+1.2\%)} & 0.338\textsubscript{(+0.1\%)} \\
    \cmidrule(lr){1-9}
    \multicolumn{2}{l}{\textit{mean}} & 0.270 & 0.288 & 0.194 & 0.249 & 0.200 & 0.104 & 0.375 \\
    \bottomrule
  \end{tabular}
  }
\end{table}
\clearpage
\section{Per-(model, prompt) delta tables}
\label{sec:delta-appendix}

The four tables below give, for every (model, prompt) configuration,
the absolute method-vs-random delta on BERTScore-F1 and RadGraph
averaged-F1, across the four transition designs and the two
aggregations \textsc{min} and \textsc{kNN}, reported for Impressions
first and Findings second.

\begin{table}[t]
  \caption{$\Delta$ BERTScore-F1 vs.\ random on \textbf{Impressions}.}
  \label{tab:delta-bertscore-impressions}
  \centering
  \footnotesize
  \setlength{\tabcolsep}{3pt}
  \resizebox{\textwidth}{!}{
  \begin{tabular}{llccccccccr}
    \toprule
    Model & Prompt & MS / m & Nov / m & DH / m & OT / m & MS / k & Nov / k & DH / k & OT / k & rand. \\
    \midrule
    Gemini-2.5-FL & P1 & $+$0.013 & $+$0.011 & $+$0.010 & $+$0.012 & $+$0.010 & $+$0.012 & $+$0.008 & $+$0.010 & 0.133 \\
    Gemini-2.5-FL & P2 & $+$0.012 & $+$0.010 & $+$0.009 & $+$0.014 & $+$0.010 & $+$0.011 & $+$0.011 & $+$0.012 & 0.150 \\
    Gemini-2.5-FL & P3 & $+$0.005 & $+$0.003 & $+$0.003 & $+$0.005 & $+$0.004 & $+$0.003 & $+$0.004 & $+$0.004 & 0.195 \\
    Gemini-3.1-FL & P1 & $+$0.005 & $+$0.005 & $+$0.010 & $+$0.006 & $+$0.004 & $+$0.004 & $+$0.010 & $+$0.009 & 0.156 \\
    Gemini-3.1-FL & P2 & $+$0.008 & $+$0.006 & $+$0.010 & $+$0.005 & $+$0.006 & $+$0.006 & $+$0.008 & $+$0.007 & 0.183 \\
    Gemini-3.1-FL & P3 & $+$0.005 & $+$0.004 & $+$0.004 & $+$0.004 & $+$0.004 & $+$0.004 & $+$0.004 & $+$0.003 & 0.201 \\
    Mistral-S-2603 & P1 & $+$0.005 & $+$0.005 & $+$0.005 & $+$0.004 & $+$0.004 & $+$0.003 & $+$0.003 & $+$0.004 & 0.131 \\
    Mistral-S-2603 & P2 & $-$0.009 & $-$0.005 & $-$0.004 & $-$0.008 & $-$0.012 & $-$0.007 & $-$0.005 & $-$0.011 & 0.163 \\
    Mistral-S-2603 & P3 & $+$0.006 & $+$0.005 & $+$0.003 & $+$0.004 & $+$0.007 & $+$0.006 & $+$0.004 & $+$0.006 & 0.200 \\
    \bottomrule
  \end{tabular}
  }
  \vspace{1pt}\par
  \scriptsize Columns: MS=\textsc{Mean-Shift}, Nov=\textsc{Novelty}, DH=\textsc{Dir-Hausdorff}, OT=\textsc{Cost-OT}. ``m''=\textsc{min} aggregation, ``k''=\textsc{kNN} aggregation. Cells are method$-$random; last column is the random baseline.
\end{table}
\begin{table}[t]
  \caption{$\Delta$ RadGraph averaged-F1 vs.\ random on \textbf{Impressions}.}
  \label{tab:delta-radgraph-impressions}
  \centering
  \footnotesize
  \setlength{\tabcolsep}{3pt}
  \resizebox{\textwidth}{!}{
  \begin{tabular}{llccccccccr}
    \toprule
    Model & Prompt & MS / m & Nov / m & DH / m & OT / m & MS / k & Nov / k & DH / k & OT / k & rand. \\
    \midrule
    Gemini-2.5-FL & P1 & $+$0.009 & $+$0.008 & $+$0.007 & $+$0.010 & $+$0.010 & $+$0.008 & $+$0.009 & $+$0.008 & 0.039 \\
    Gemini-2.5-FL & P2 & $+$0.013 & $+$0.011 & $+$0.009 & $+$0.015 & $+$0.012 & $+$0.013 & $+$0.012 & $+$0.013 & 0.056 \\
    Gemini-2.5-FL & P3 & $+$0.005 & $+$0.001 & $+$0.001 & $+$0.004 & $+$0.004 & $+$0.002 & $+$0.003 & $+$0.003 & 0.077 \\
    Gemini-3.1-FL & P1 & $+$0.005 & $+$0.003 & $+$0.005 & $+$0.006 & $+$0.005 & $+$0.003 & $+$0.005 & $+$0.007 & 0.052 \\
    Gemini-3.1-FL & P2 & $+$0.013 & $+$0.007 & $+$0.009 & $+$0.007 & $+$0.012 & $+$0.009 & $+$0.008 & $+$0.011 & 0.072 \\
    Gemini-3.1-FL & P3 & $+$0.005 & $+$0.002 & $+$0.002 & $+$0.004 & $+$0.004 & $+$0.004 & $+$0.003 & $+$0.003 & 0.079 \\
    Mistral-S-2603 & P1 & $+$0.000 & $+$0.001 & $+$0.001 & $+$0.000 & $-$0.000 & $-$0.000 & $-$0.000 & $-$0.001 & 0.017 \\
    Mistral-S-2603 & P2 & $+$0.012 & $+$0.016 & $+$0.015 & $+$0.012 & $+$0.011 & $+$0.014 & $+$0.013 & $+$0.012 & 0.049 \\
    Mistral-S-2603 & P3 & $+$0.005 & $+$0.003 & $+$0.000 & $+$0.003 & $+$0.005 & $+$0.004 & $+$0.001 & $+$0.004 & 0.071 \\
    \bottomrule
  \end{tabular}
  }
  \vspace{1pt}\par
  \scriptsize Columns: MS=\textsc{Mean-Shift}, Nov=\textsc{Novelty}, DH=\textsc{Dir-Hausdorff}, OT=\textsc{Cost-OT}. ``m''=\textsc{min} aggregation, ``k''=\textsc{kNN} aggregation. Cells are method$-$random; last column is the random baseline.
\end{table}
\begin{table}[t]
  \caption{$\Delta$ BERTScore-F1 vs.\ random on \textbf{Findings}, per (model, prompt).}
  \label{tab:delta-bertscore-findings}
  \centering
  \footnotesize
  \setlength{\tabcolsep}{3pt}
  \resizebox{\textwidth}{!}{
  \begin{tabular}{llccccccccr}
    \toprule
    Model & Prompt & MS / m & Nov / m & DH / m & OT / m & MS / k & Nov / k & DH / k & OT / k & rand. \\
    \midrule
    Gemini-2.5-FL & P1 & $+$0.005 & $+$0.005 & $+$0.003 & $+$0.007 & $+$0.007 & $+$0.004 & $+$0.002 & $+$0.006 & 0.147 \\
    Gemini-2.5-FL & P2 & $+$0.008 & $+$0.006 & $+$0.006 & $+$0.007 & $+$0.006 & $+$0.004 & $+$0.007 & $+$0.006 & 0.182 \\
    Gemini-2.5-FL & P3 & $+$0.004 & $+$0.002 & $+$0.002 & $+$0.003 & $+$0.004 & $+$0.002 & $+$0.001 & $+$0.003 & 0.223 \\
    Gemini-3.1-FL & P1 & $+$0.002 & $-$0.000 & $+$0.002 & $+$0.002 & $+$0.001 & $-$0.000 & $+$0.003 & $+$0.002 & 0.188 \\
    Gemini-3.1-FL & P2 & $+$0.004 & $+$0.005 & $+$0.004 & $+$0.004 & $+$0.005 & $+$0.004 & $+$0.004 & $+$0.006 & 0.212 \\
    Gemini-3.1-FL & P3 & $+$0.001 & $+$0.001 & $-$0.000 & $-$0.000 & $+$0.000 & $+$0.001 & $-$0.000 & $-$0.001 & 0.227 \\
    Mistral-S-2603 & P1 & $+$0.012 & $+$0.008 & $+$0.012 & $+$0.013 & $+$0.012 & $+$0.008 & $+$0.012 & $+$0.013 & 0.160 \\
    Mistral-S-2603 & P2 & $+$0.007 & $+$0.007 & $+$0.006 & $+$0.008 & $+$0.008 & $+$0.006 & $+$0.008 & $+$0.007 & 0.200 \\
    Mistral-S-2603 & P3 & $-$0.000 & $+$0.000 & $+$0.000 & $-$0.000 & $-$0.001 & $-$0.000 & $+$0.000 & $-$0.000 & 0.228 \\
    \bottomrule
  \end{tabular}
  }
  \vspace{1pt}\par
  \scriptsize Columns: MS=\textsc{Mean-Shift}, Nov=\textsc{Novelty}, DH=\textsc{Dir-Hausdorff}, OT=\textsc{Cost-OT}. ``m''=\textsc{min} aggregation, ``k''=\textsc{kNN} aggregation. Cells are method$-$random; last column is the random baseline.
\end{table}
\begin{table}[t]
  \caption{$\Delta$ RadGraph averaged-F1 vs.\ random on \textbf{Findings}.}
  \label{tab:delta-radgraph-findings}
  \centering
  \footnotesize
  \setlength{\tabcolsep}{3pt}
  \resizebox{\textwidth}{!}{
  \begin{tabular}{llccccccccr}
    \toprule
    Model & Prompt & MS / m & Nov / m & DH / m & OT / m & MS / k & Nov / k & DH / k & OT / k & rand. \\
    \midrule
    Gemini-2.5-FL & P1 & $+$0.006 & $+$0.006 & $+$0.004 & $+$0.008 & $+$0.008 & $+$0.004 & $+$0.004 & $+$0.008 & 0.070 \\
    Gemini-2.5-FL & P2 & $+$0.008 & $+$0.005 & $+$0.004 & $+$0.008 & $+$0.006 & $+$0.003 & $+$0.004 & $+$0.007 & 0.092 \\
    Gemini-2.5-FL & P3 & $+$0.002 & $+$0.000 & $+$0.001 & $+$0.004 & $+$0.004 & $+$0.001 & $+$0.000 & $+$0.004 & 0.128 \\
    Gemini-3.1-FL & P1 & $+$0.000 & $-$0.002 & $+$0.001 & $+$0.001 & $+$0.001 & $-$0.001 & $+$0.001 & $+$0.001 & 0.097 \\
    Gemini-3.1-FL & P2 & $+$0.010 & $+$0.006 & $+$0.004 & $+$0.008 & $+$0.009 & $+$0.005 & $+$0.005 & $+$0.009 & 0.117 \\
    Gemini-3.1-FL & P3 & $-$0.002 & $-$0.002 & $-$0.003 & $-$0.002 & $-$0.002 & $-$0.002 & $-$0.003 & $-$0.003 & 0.126 \\
    Mistral-S-2603 & P1 & $+$0.003 & $+$0.004 & $+$0.005 & $+$0.004 & $+$0.002 & $+$0.003 & $+$0.003 & $+$0.003 & 0.052 \\
    Mistral-S-2603 & P2 & $+$0.014 & $+$0.016 & $+$0.011 & $+$0.013 & $+$0.014 & $+$0.013 & $+$0.013 & $+$0.013 & 0.096 \\
    Mistral-S-2603 & P3 & $-$0.001 & $+$0.002 & $+$0.001 & $-$0.001 & $-$0.001 & $+$0.001 & $-$0.000 & $-$0.000 & 0.129 \\
    \bottomrule
  \end{tabular}
  }
  \vspace{1pt}\par
  \scriptsize Columns: MS=\textsc{Mean-Shift}, Nov=\textsc{Novelty}, DH=\textsc{Dir-Hausdorff}, OT=\textsc{Cost-OT}. ``m''=\textsc{min} aggregation, ``k''=\textsc{kNN} aggregation. Cells are method$-$random; last column is the random baseline.
\end{table}

\end{document}